\documentclass[runningheads]{llncs}
\usepackage{graphicx}
\usepackage{enumitem}
\usepackage{todonotes}
\usepackage{comment}
\usepackage{xspace}
\newcommand\sysname{D-Rax\xspace}
\usepackage{MnSymbol}
\usepackage[subtle,margins=normal,leading=normal]{savetrees}
\begin{document}
\title{D-Rax: Domain-specific Radiologic assistant leveraging multi-modal data and eXpert model predictions}
\titlerunning{D-Rax}





\author{Hareem Nisar$^{1}$, Syed Muhammad Anwar $^{1,2}$, Zhifan Jiang$^{1}$,\\Abhijeet Parida $^{1,3}$, Ramon Sanchez-Jacob$^{1,2}$, Vishwesh Nath $^4$,
\\Holger R. Roth$^4$, Marius George Linguraru $^{1,2}$}
\authorrunning{H. Nisar et al.}
\institute{ Children's National Hospital, Washington, DC, USA \and George Washington University, Washington, DC, USA \and Universidad Politécnica de Madrid, Madrid, Spain \and Nvidia Corporation, Santa Clara, CA, USA}
\maketitle

\begin{abstract}
Large vision language models (VLMs) have progressed incredibly from research to applicability for general-purpose use cases. LLaVA-Med, a pioneering large language and vision assistant for biomedicine, can perform multi-modal biomedical image and data analysis to provide a natural language interface for radiologists. While it is highly generalizable and works with multi-modal data, it is currently limited by well-known challenges that exist in the large language model space. Hallucinations and imprecision in responses can lead to misdiagnosis which currently hinder the clinical adaptability of VLMs. To create precise, user-friendly models in healthcare, we propose \sysname – a domain-specific, conversational, radiologic assistance tool that can be used to gain insights about a particular radiologic image. In this study, we enhance the conversational analysis of chest X-ray (CXR) images to support radiological reporting, offering comprehensive insights from medical imaging and aiding in the formulation of accurate diagnosis. \sysname is achieved by fine-tuning the LLaVA-Med architecture on our curated enhanced instruction-following data, comprising of images, instructions, as well as disease diagnosis and demographic predictions derived from MIMIC-CXR imaging data, CXR-related visual question answer (VQA) pairs, and predictive outcomes from multiple expert AI models. We observe statistically significant improvement in responses when evaluated for both open and close-ended conversations.    
Leveraging the power of state-of-the-art diagnostic models combined with VLMs, \sysname empowers clinicians to interact with medical images using natural language, which could potentially streamline their decision-making process, enhance diagnostic accuracy, and conserve their time.

\keywords{Large vision language models  \and Radiologic assistant \and Chest X-ray \and Expert models.}
\end{abstract}

\section{Introduction}

Burnout in radiology is on the rise globally leading to chronic job dissatisfaction and critical under-staffing~\cite{Fawzy2023}. Radiologists routinely spend extensive time meticulously analyzing medical images to identify pathologies and diagnose diseases, which is vital in guiding treatment decisions and ensuring appropriate patient care. The retrospective error rate among radiologic exams has been reported to be around 30\%~\cite{Cindy2013}. Cindy et al.~ \cite{Cindy2013} assess these errors to be either cognitive, like false initial assessment, framing bias (i.e., misinterpretation caused by choice of words), and premature closure of a case, or system-related errors, such as long working shifts, repetitive tasks, and lighting conditions. Many of these factors contribute to visual and mental fatigue for radiologists, further contributing to misdiagnosis and poor patient outcomes. Another challenge is miscommunication between radiologists, clinicians, and patients, often caused by inefficient reporting.

With the constant increase in workload in radiology departments~\cite{Bruls2020}, generative artificial intelligence (AI) can play a crucial role in reducing the burden and improving healthcare~\cite{Mukherjee2023}.
Recent large vision language models (VLMs) such as LLaVA-Med~\cite{LLaVA-Med} have been created to assist clinicians in interpreting complex medical imaging and provide visual question answering (VQA) in natural language settings. Despite its enhanced capabilities for medical image analysis and interpretation, LLaVA-Med is highly generalized and cannot precisely answer specific questions~\cite{Wu2023}, as well as suffers from hallucinations that can result in misdiagnosis. Another challenge in the integration and adoption of AI-driven technologies among healthcare professionals is user-friendliness of the tool~\cite{Hemmer2022}.
These clinical and technological challenges necessitate a “Radiology Assistant” tool that can facilitate report writing and provide a natural-language interface to discuss imaging features, pathological findings, and disease diagnosis with the radiologist. 

To address these challenges, we propose a novel, domain-specific VLM, called \sysname, which empowers radiologists to interact with images using natural language prompts and questions, similar to how they converse with colleagues. 
Furthermore, our model is equipped with the knowledge of identifying pathologies and diagnostic reasoning. \sysname leverages established AI models~\cite{densenet} to incorporate expert model diagnostic predictions for multiple diseases, thus reducing the risk of missed findings and aiding in achieving more accurate diagnoses. To exemplify the utility of our domain-specialist VLM, we chose chest X-ray (CXR) images for this study. CXRs are among the most commonly performed imaging studies and play a crucial role in the diagnosis and management of a wide range of medical conditions, including respiratory diseases, cardiac abnormalities, and thoracic injuries. 
The novelty and contributions of this work can be summarized as: 
\begin{itemize}[label=\raisebox{0.25ex}{\tiny$\bullet$}]
    \item \textit{Enhanced instruction-following training with expert model predictions.} We introduce a novel, domain-specific, and multi-modal instruction-following training strategy enriched with multiple expert model predictions for large VLMs. 
    \item \textit{Expert-enhanced instruction-following data generation.} We use MIMIC-CXR and Medical-Diff-VQA datasets to generate baseline instruction-following data for the design of conversational image analysis tools. State-of-the-art (SOTA) AI (expert) models are incorporated to add diagnostic and demographic predictions to the baseline, thus creating expert-enhanced visual instruction-following training data.
    \item \textit{\sysname.} Our expert-enhanced instruction-following training leads to a more accurate radiologic assistance tool, demonstrated by comprehensive comparisons. The same training paradigm can potentially benefit other conversational AI tools.
\end{itemize}

\section{Related Work}
The introduction of foundational large VLMs has flooded the gates for the design of complex multi-modal AI tools. Flamingo~\cite{Alayrac2022} is one of the earliest multi-modal VLMs that bridged the gap between image-only and text-only methods. It combines prompts and multi-line chains of thought to produce sensible outcomes. Another notable example is the Large Language and Vision Assistant (LLaVA)~\cite{llava} model that leverages a multi-modal architecture capable of processing both visual and textual information. Both of these VLM frameworks closely follow the technicalities from the Contrastive Language-Image Pre-training (CLIP)~\cite{Radford2021} model, which is a technique to associate images with corresponding textual descriptions. Such VLMs are widely adopted in the computer vision industry and are a gateway to many advances in biomedicine.

In the realm of biomedical VLMs, BioMedClip~\cite{Zhang2023} is an important foundation model, with vision-language processing capabilities, enabling several standard biomedical imaging tasks such as classification and visual question-answering. LLaVA-Med~\cite{LLaVA-Med}, a specialized version of LLaVA, is tailored for biomedical applications, including radiology, to enable clinicians to interact with medical images in a conversational language setting, thereby facilitating more efficient radiological workflows. 
OphGLM~\cite{gao2023ophglm} combined expert model deductions with large language models (LLM) by generating a diagnostic report from retinal images. Most biomedical VLMs, however, are generalized and suffer from hallucinations, inaccurate diagnosis, and imprecise question answering. A domain-specialized tool in radiology can help overcome these challenges and provide accurate outcomes.



\section{Methods}






\subsection{Data}
\label{subsec:data}
\paragraph{Baseline Instruction-following Data}
The multi-modal nature of our task requires both vision and language information. In this study, we use the MIMIC-CXR and Medical-Diff-VQA datasets to generate a baseline instruction-following dataset for our experiments. MIMIC-CXR~\cite{mimic2,mimic3,mimic1} is a large open-access dataset of $377,110$ CXRs with structured labels on cardiopulmonary conditions derived from $227,827$ free-text radiology reports. 
Medical-Diff-VQA~\cite{mdvqa1,mdvqa2,mimic3} is a derivative of the MIMIC-CXR dataset containing $700,703$ question-answer (QA) pairs derived from CXRs. The questions are divided into seven categories: abnormality, presence, view, location, level, type, and difference. Each category can hold either open-ended questions such as ‘why, what, how’, etc. with dynamic natural language answers or close-ended questions such as ‘Is there’ with binary answers like ‘yes/no’. To limit the complexity of the evaluation, we did not focus on longitudinal changes, therefore the difference QAs were removed from the current evaluation. As a result, only a single image per patient was extracted to form the test set. 
Table~\ref{tab:data} summarizes the data distribution for the baseline dataset.
\begin{table}[htbp]
\caption{Baseline instruction-following data - Summary of train and test datasets and percentage distribution of QA categories.}
\label{tab:data}
\centering
\begin{tabular}{l|l|ccccccc}
\hline
    &  & \textbf{Total} & $\%$Abnormality & Presence & View & Location & Level & Type \\
    \hline
    \textbf{Train} & \#QA Pairs & $429,000$ & $27.1$ & $29.1$ & $10.5$ & $15.7$ & $12.5$ & $5.1$ \\
    $129,232$ & \#Open & $219,305$ & $24.6$ & $0$ & $10.2$ & $30.6$ & $24.5$ & $10.1$ \\
    images & \#Close & $209,695$ & $29.8$ & $59.4$ & $10.8$ & $0$ & $0$ & $0$ \\
    \hline 
    \textbf{Test} & \#QA Pairs & $13,688$ & $26.8$ & $29.3$ & $13.5$ & $14$ & $11.6$ & $4.8$ \\
    $4,190$ & \#Open & $6,683$ & $23.6$ & $0$ & $14$ & $28.8$ & $23.7$ & $9.9$ \\
    images & \#Close & $7,005$ & $29.8$ & $57.2$ & $13$ & $0$ & $0$ & $0$ \\
    \hline 
\end{tabular}
\end{table}

\paragraph{Enhanced Expert Instruction-following Data}
We enhanced the baseline dataset by incorporating MIMIC-CXR along with QA conversations and integrating expert model predictions using pre-trained models from the TorchXRayVision~\cite{torchxray} model zoo. Expert predictions on the MIMIC-CXR dataset fall into one of the four categories - diseases, age, race, and view (Table~\ref{tab:data2}). The outcomes of these SOTA AI model predictions are appended to the baseline dataset to create our expert-model enhanced instruction-following dataset. The medical conditions in the first category include cardiomegaly, atelectasis, pneumonia, infiltration, fracture, enlarged cardio mediastinum, lung opacity, pneumothorax, emphysema, hernia, lung lesion, pleural thickening, edema, effusion, fibrosis, nodule, mass, and consolidation.

\begin{table}[!htbp]
\caption{Expert-model enhanced instruction-following data - Details on the AI model, training dataset, and labels used for each category.}
\label{tab:data2}
\centering
\resizebox{0.90\textwidth}{!}{\begin{tabular}{llll}
\hline
    Expert Predictions & Model & Dataset & Labels \\
    \hline
   \textbf{Disease diagnosis} & DenseNet121~\cite{densenet} & MIMIC-CXR & CheXpert\cite{chexpert} \\
    \textbf{Patient age} & Regression~\cite{Ieki2022} & NIH ChestX-ray8~\cite{nih} & \\
    \textbf{Patient race} & Classifier~\cite{Gichoya2022} & MIMIC-CXR & Asian, Black, White\\
    \textbf{CXR view position} & ChestViewSplit~\cite{cvsplit} & & Frontal, Lateral \\
    \hline 
\end{tabular}
}
\end{table}

\subsection{Domain Specific Radiologic Assistant Design}

\begin{figure}[!htbp]
\begin{center}
\includegraphics[width=0.9\textwidth]{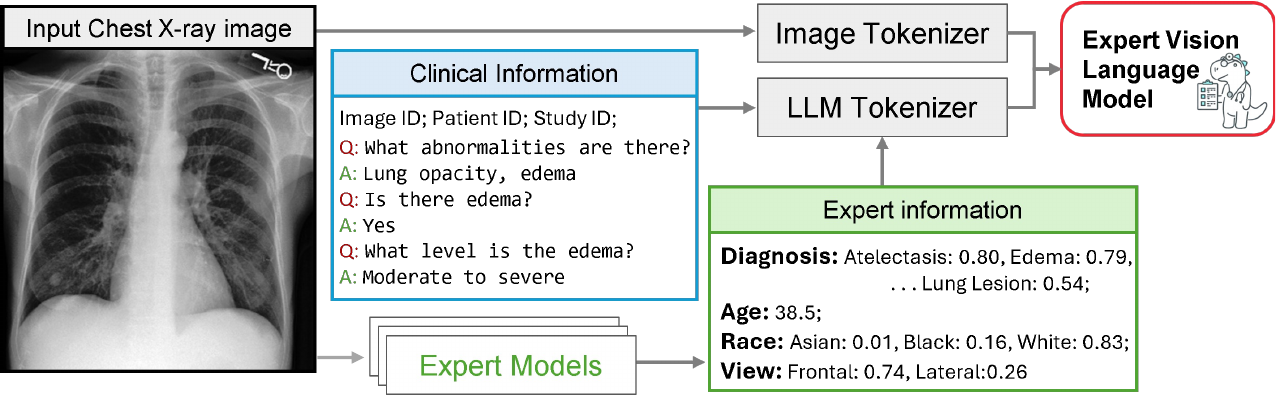}
\end{center}
\caption{
\label{fig:workflow} Overview of our expert vision language model \sysname design - Training data includes multimodal data including visual information (Chest X-ray images) and textual information (VQA from radiology reports, and expert model predictions).}
\end{figure}

The original LLaVA-Med model was trained on 15 million figure-caption pairs from PubMed~\cite{Zhang2023}. While this teaches the model the context of biomedical application, we argue that for the sensitive process of medical imaging diagnosis, it is beneficial to develop a domain-specific VLM. Therefore, we perform end-to-end instruction tuning by training our model with CXRs and VQA-derived instructions generated from the associated radiology reports. In the process, we generated novel and enhanced instruction-following data for CXRs by incorporating predictions from expert models (Figure~\ref{fig:workflow}). 

\textit{Network Architecture:} The definition of the expert VLM model follows the network architecture proposed in~\cite{llava}. We chose Llama2~\cite{llama} as our LLM due to the availability of the pre-trained checkpoints and particularly used the Llama2-7B model. 
The visual encoder was kept consistent as ViT-Large/14 which is a pre-trained CLIP model. For any given input image $X_v$ and a series of question and answer defined as $(X_q^1, X_a^1, X_q^2, X_a^2 ... X_q^T, X_a^T)$. First, $X_v$ is transformed into a set of visual features $Z_v$ by the CLIP model. For training with the instruction tuning data, the visual encoder is kept frozen and a trainable projection matrix $W$ is used to convert the visual features $Z_v$ into language embedding tokens $H_v = W.Z_v$ that can be jointly used with the language embedding tokens $H_q$ of the questions $X_q$. The output of the model is $X_a$. Please note that $H_q$ and $H_v$ are used as inputs to the LLM model of the entire framework. For VLM training, we used the LLaVA-v1.5-7B~\cite{llava} model as a baseline and the model weights were initialized with Llama2-7B weights added with a delta from LLaVA training. 

\textit{Expert Model Enhanced Instruction Tuning Data:} Within the Medical-Diff-VQA data, for any given input image $X_v$, there exists a multi-turn conversation that pertains to different categories of questions related to abnormality, presence, view, location, level, and type  
such that $X_q$ belongs to a specific category. The dataset was enhanced with expert response $X_e$ to the questions which when updated reads as $(\{X_e, X_q^1\}:X_a^1, \{X_e, X_q^2\}:X_a^2, ... \{X_e, X_q^T\}:X_a^T)$. Medical-Diff-VQA also provides QA pairs for difference (with a reference image), which we have not used in our experiments.  

\textit{End-to-end Training:} The complete training of domain-specific VLMs (particularly LLaVA) involves two steps: (1) concept alignment to biomedical concepts from large data from PubMed including figures in published articles, captions, and inline references to figures, and (2) an instruction tuning step, where both the projection layer and the language model are updated. In our method, we perform the instruction fine tuning with the multi-modal, expert-enhanced dataset for CXRs presented in Section~\ref{subsec:data}. We also perform a set of experiments to establish the usefulness of employing expert model predictions to guide the radiology assistant's answers. Finally, we evaluate the performance of open- and close-ended conversational questions to establish the efficacy of our proposed strategy.

\subsection{Experiments}

For visual question answers related to radiology, LLaVA-Med was finetuned and evaluated on the VQA-RAD and SLAKE datasets. However, the data used covers multiple modalities and is relatively small in size, for instance: VAQ-RAD~\cite{VQARAD} has 315 radiology images and 3,515 QA pairs, and SLAKE~\cite{slake} has 642 images and 7,000 QA pairs. For \sysname, we utilize MIMIC, one of the largest domain-specific medical imaging data, and the associated VQA pairs. We further leverage expert models to provide more context and expert knowledge of the language model. 
We hypothesize that with expert model prediction, the trained radiologic assistant will have better outcomes in terms of reducing hallucinations and providing more precise and correct responses. 

To establish the efficacy of utilizing model predictions related to abnormality, age, race, and view, we ran multiple experiments for end-to-end instruction tuning of various LLaVA models. In particular, we use the following pre-trained models: LLaVA, LLaVA-Med finetuned on VQA-RAD (LLaVA-Med-RAD), and LLaVA-Med finetuned on SLAKE (LLaVA-Med-SLAKE). VQA-RAD and SLAKE are selected since they are closely related to our research question, however, the data represented has a much larger scope with fewer examples to enable the development of precise models. Overall we performed six different experiments, including end-to-end instruction fine-tuning with a model initialized with weights from the three aforementioned pre-trained models. For each of these model initializations, the instruction fine-tuning was performed both for the baseline dataset (images and VQA-derived instructions from MIMIC) and an enhanced dataset with augmentation of expert model predictions. The training was performed for a single epoch, with a learning rate of $2e^{-5}$ and an effective batch size of $8$. 



\subsection{Evaluation}
For performance evaluation, two metrics were utilized: accuracy and token recall, depending on the type of questions evaluated. For close-ended questions, the task can be considered as a classification, and hence we used accuracy. For open-ended questions, token recall measures the ratio of tokens correctly generated by the trained model according to the ground truth. Evaluating VLMs, particularly for open-ended questions is still a difficult problem and some approaches try to use OpenAI's GPT-4 to evaluate the similarity between ground truth and predicted answers~\cite{LLaVA-Med,llava}. The inference of the finetuned model required 20G of GPU memory and could generate answers for $10,000$ questions per hour on a single NVIDIA H100 80G GPU. 


\section{Results}

\begin{figure}[!htbp]
\begin{center}
\includegraphics[width=0.95\textwidth]{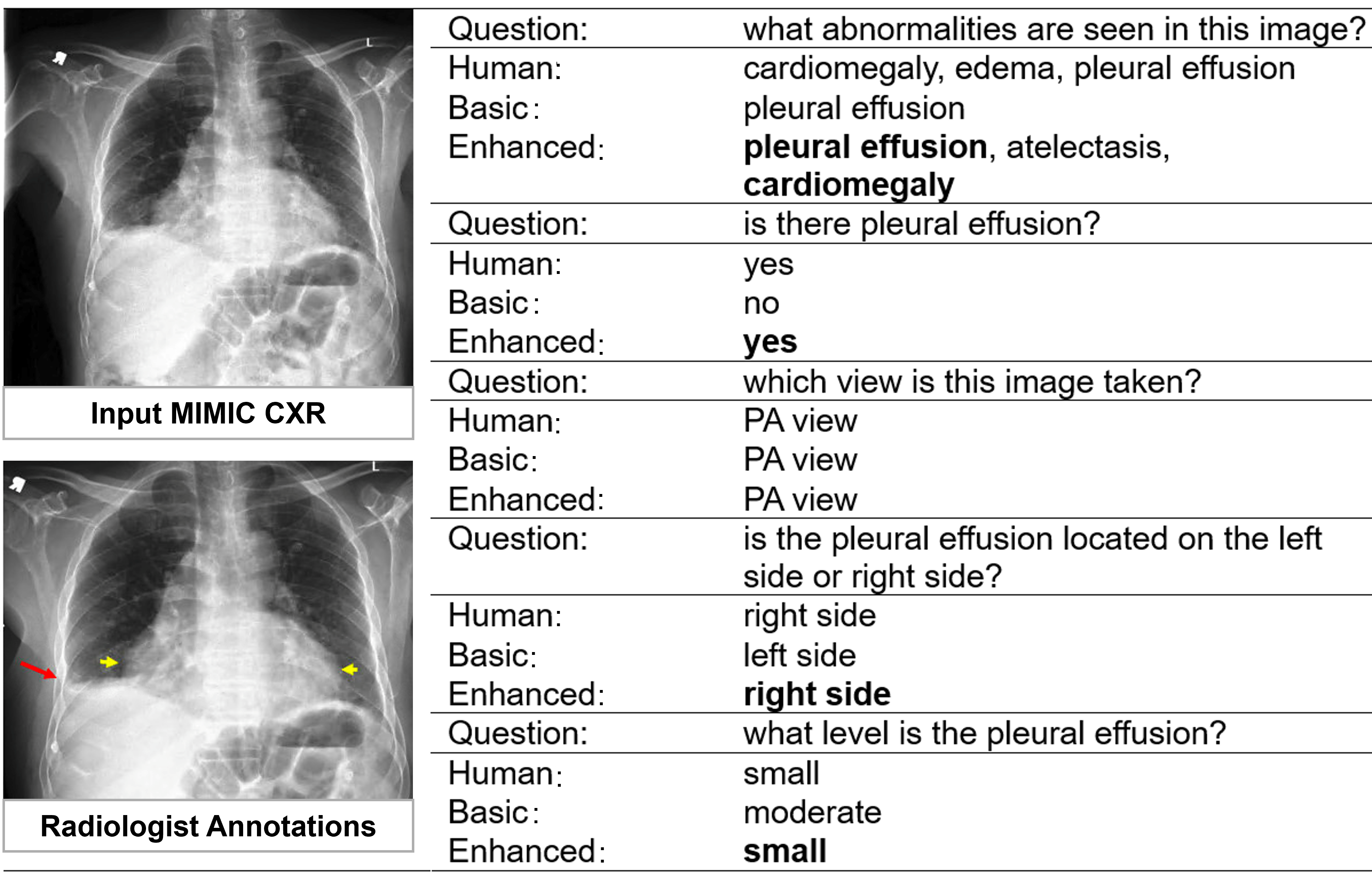}
\end{center}
\caption{Qualitative evaluation: conversations provided by VLMs trained on basic and expert enhanced data. The red arrow shows the area of the pleural effusion and the yellow arrows outline the lateral margins of the enlarged heart (cardiomegaly) provided by the radiologist, which were correctly identified by \sysname.}
\label{fig:result}
\end{figure}

\paragraph{Performance of Enhanced Instruction}
Figure~\ref{fig:result} shows the qualitative evaluation of \sysname by showing an example of conversations on a given CXR, as generated by VLMs trained on basic and expert-enhanced data.
The results from quantitative evaluation (Table~\ref{tab:res}) indicates that
the enhanced expert instruction training allows for statistically significant improvements in the model performance for abnormality and presence questions (both open and closed-ended). Meanwhile, for location, level, and type questions, where the expert model provides no explicit information, training on both basic and enhanced data mostly yields similar performance and even showcases improvements when using the LLaVA-Med-RAD model as the base. Intriguingly, in addressing the view questions, the expert model introduces different view information but does not affect the model's capacity to derive correct answers from images and questions. Overall, expert model-enhanced instruction training enables higher performance without impeding the pre-trained model's inherent ability to comprehend queries and images.

\begin{table}[!htbp]
\caption{Quantitative evaluation: token recall (\%) for open-ended questions (O) and accuracy (\%) for close-ended questions (C) are reported to show the performance of domain-specific VLM with basic and enhanced instruction tuning strategies across various question types. Each value is an average and standard deviation of three inferences. 
The asterisks show statistical significance across paired comparisons using the Wilcoxon signed rank test 
(* for p-value $< 0.05$ and ** for p-value $< 0.001$).}
\label{tab:res}
\centering
\begin{tabular}{ll|ll|ll|ll}
\hline
  \multicolumn{2}{l|}{Metrics (\%)} &
  \multicolumn{2}{c|}{LLaVA} &
  \multicolumn{2}{c|}{LLaVA-Med-RAD} & 
  \multicolumn{2}{c}{LLaVA-Med-SLAKE} \\
  \hline
    \multicolumn{2}{l|}{Question Type} & Basic & Enhanced & Basic & Enhanced & Basic & Enhanced \\ \hline
    Abnormality & (O) & $40.6 (0.5) $ & $\mathbf{41.7 (0.3)}$ & $39.8 (0.5) $ & $\mathbf{41.7 (0.1)}^{*}$ & $39.5 (0.5) $ & $\mathbf{42.0 (0.7)}^{**}$ \\
    & (C) & $70.1 (1.3)$ & $\mathbf{71.5 (0.3)}^{*}$ & $70.3 (0.9)$ & $\mathbf{72.8 (0.6)}^{**}$ & $68.9 (0.4)$ & $\mathbf{71.8 (1.1)}^{**}$ \\ 

    Presence & (C) & $76.1 (0.2)$ & $\mathbf{77.7 (0.1)}^{**}$ & $75.5 (0.2)$ & $\mathbf{77.6 (0.4)}^{**}$ & $75.0 (0.3)$ & $\mathbf{77.9 (0.4)}^{**}$ \\ 

    View & (O) & $99.7 (0.0)$ & $99.7 (0.0)$ & $99.6 (0.0)$ & $99.6 (0.0)$ & $99.6 (0.1)$ & $99.6 (0.1)$ \\ 
    & (C) & $99.0 (0.2)$ & $98.8 (0.1)$ & $98.9 (0.2)$ & $99.1 (0.2)$ 
    & $98.8 (0.1)$ & $98.6 (0.2)$ \\ 

    Location & (O) & $61.8 (0.0)$ & $61.6 (0.4)$ & $60.2 (0.4)$ & $\mathbf{61.6 (0.6)}^{*}$ & $60.3 (0.2)$ & $\mathbf{61.8 (0.5)}^{*}$ \\ 

    Level & (O) & $59.1 (0.8)$ & $59.5 (0.4)$ & $58.8 (0.5)$ & $\mathbf{60.4 (0.4)}^{*}$ & $59.2 (0.8)$ & $60.0 (0.9)$ \\ 

    Type & (O) & $60.6 (1.0)$ & $60.6 (0.8)$ & $58.9 (1.0)$ & $58.5 (1.0)$ & $58.1 (0.2)$ & $58.4 (1.3)$ \\ 
    \hline
    
    Average & (O) & $61.3$ & $61.6$ & $60.4$ & $\mathbf{61.6}^{**}$ & $60.4$ & $\mathbf{61.7}^{**}$\\
    & (C) & $77.3$ & $\mathbf{78.6}^{**}$ & $77.0$ & $\mathbf{79.0}^{**}$ & $76.3$ & $\mathbf{78.8}^{**}$ \\
    \hline
    
\end{tabular}
\end{table}

\paragraph{Comparison with Expert Models}
D-Rax is not expected to outperform disease-specific expert models that are restricted to answering simple and close-ended (C) questions based on classification. Analysis of experiments on Abnormality (C) questions shows that the diagnostic accuracy of expert models of \textbf{70.4\%} was comparable to VLMs (p-value $> 0.08$), except in 1/18 inferences where enhanced LLaVA-Med-RAD (Table~\ref{tab:res}) outperformed significantly the expert models (p-value $= 0.01$). However, identifying abnormalities is just one aspect of the VLM. VLMs can handle complex and nuanced questions, unlike expert models which cannot understand natural language queries.

\paragraph{Ablation Studies}
While we maintained the test set's characteristics by extracting one image per patient (Section~\ref{subsec:data}), the following ablation study (Table~\ref{tab:ab1}) shows the results evaluated on an extended test set, including all the images from each patient. The improved results demonstrate the robustness of the method when tested on larger data. However, since evaluation of the larger test data is computationally expensive, most results reported in the paper are on the smaller test set.

\begin{table}[thpb]
\caption{Evaluation on an extended test set. 
The asterisks show statistical significance across paired comparisons using the Wilcoxon signed rank test 
(* for p-value $< 0.05$ and ** for p-value $< 0.001$).}
\label{tab:ab1}
\centering
\resizebox{0.85\textwidth}{!}{\begin{tabular}{l|cc|c|cc|c|c|c|cc}
    \hline
    Question Type & 
    \multicolumn{2}{c|}{Abnormality} &
    Presence & 
    \multicolumn{2}{c|}{View} &
    Location &
    Level &
    Type & 
    \multicolumn{2}{c}{Average} \\
    Metrics (\%)& 
    (O) &
    (C) & 
    (C) & 
    (O) &
    (C) & 
    (O) &
    (O) & 
    (O) & 
    (O) &
    (C) \\ 
    \hline
    \multicolumn{9}{c}{\textbf{Test Set} $4,190$ images $13,688$ QA pairs} \\
    \hline 
    LLaVA-Basic & $40.6$ & $70.1$ & $76.1$ & 
    $99.7$ & $99.0$ & $61.8$ & $59.1$ & $60.6$ & $61.3$ & $77.3$\\
    LLaVA-Enhanced & $\mathbf{41.7}$ & $\mathbf{71.5}^{*}$ & $\mathbf{77.7}^{**}$ & $99.7$ & $98.8$ & $61.6$ & $59.5$ & $60.6$ 
    & $61.6$ & $\mathbf{78.6}^{**}$\\
    \hline
    \multicolumn{9}{c}{\textbf{Extended Test Set} $32,205$ images $107,379$ QA pairs} \\
    \hline 
    LLaVA-Basic & $42.7$ & $73.4$ & $76.1$ & 
    $99.5$ & $98.7$ & $61.8$ & $57.7$ & $60.0$ & $59.9$ & $77.7$ \\
    LLaVA-Enhanced & $\mathbf{43.9}^{*}$ & $\mathbf{75.4}^{**}$ & $\mathbf{77.2}^{**}$ & 
    $99.5$ & $98.7$ & $61.9$ & $\mathbf{58.7}^{*}$ & $59.9$ & $60.4^{*}$ & $\mathbf{78.9}^{**}$ \\
    \hline
\end{tabular}
}
\end{table}

\section{Discussion and Conclusion}

Our goal for developing \sysname, a domain-specific expert model-guided radiologic assistant, is to reduce the hallucinations and improve the precision observed in responses from VLMs.
We achieve this goal by establishing a novel training paradigm incorporating predictions from expert models.
Hence, in our target application of CXR analysis, we embed expert predictions for disease, age, race, and view with the VQA instructions generated from radiological reports. 
Our results validate our hypothesis that (1) domain-specific knowledge, such as the use of MIMIC-CXR and Medical-Diff-VQA for CXR analysis, extracted from clinical radiology reports introduces a human factor into the model resulting in reduced hallucinations and allowing the system to provide precise information; and (2) addition of expert information from SOTA AI models generates statistically significant improved outcomes, enhancing accuracy of answering both open and close-ended questions in a conversation. \sysname has the potential to enable a natural flow of diagnostic reasoning, enhance communication among clinicians, provide clear and accessible information to patients, and ultimately improve clinical care.





\newpage
\bibliographystyle{splncs04}
\bibliography{Ref}


\newpage
\appendix
\section{Expert Enhanced Training}\label{app:data-splits}
\begin{figure}[h]
\begin{center}
\includegraphics[width=0.5\textwidth]{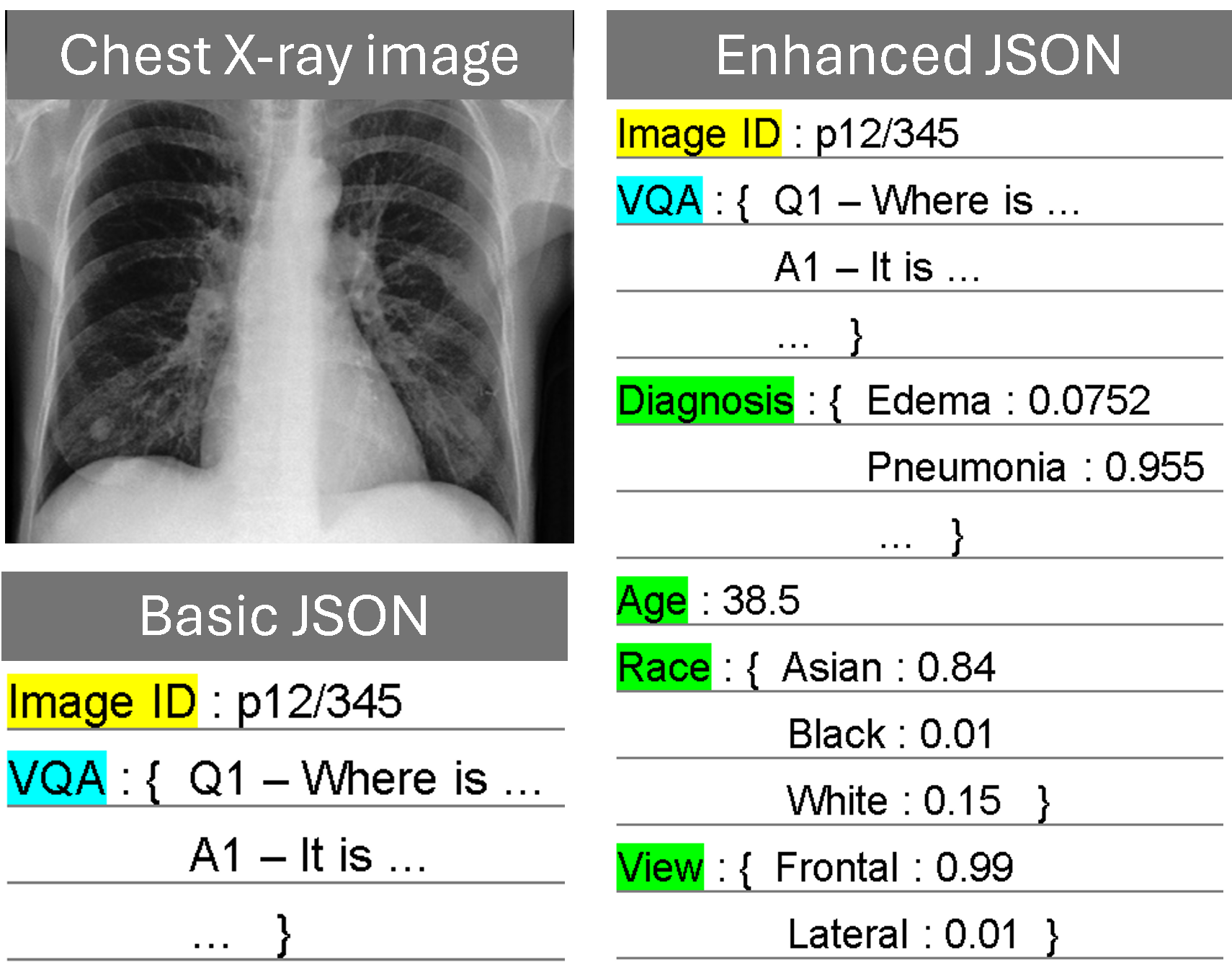}
\end{center}
\caption{
\label{fig:appendix} Data organization for expert enhanced training containing the following information: (1) image identifiers, (2) question-answer pairs, (3) diagnostic prediction on 18 medical conditions, (4) predicted age of the patient, (5) predicted race of the patient, and (6) predicted view of the CXR.}
\end{figure}

\section{No Abnormality Questions}\label{app:no-abnormal}
\begin{table}[h]
\caption{Removing abnormality questions ($27\%$ of the data) from training. Token recall (\%) for open-ended questions (O) and accuracy (\%) for close-ended questions (C) are reported to show the performance of LLaVA models finetuned on enhanced instruction dataset using $100\%$ and $73\%$ data, respectively. Each value is an average of three inferences and standard deviations are reported in parentheses. The asterisks show statistical significance across paired comparisons using the Wilcoxon signed rank test 
(* for p-value $< 0.05$ and ** for p-value $< 0.001$).}
\label{tab:supp1}
\centering
\begin{tabular}{ll|ll|ll}
\hline
  \multicolumn{2}{l|}{Metrics (\%)} &
  \multicolumn{2}{c|}{LLaVA-Enhanced} &
  \multicolumn{2}{c}{LLaVA-Med-RAD-Enhanced} \\
  \hline
    \multicolumn{2}{l|}{Question Type} & $100\%$ data & $73\%$ data & $100\%$ data & $73\%$ data \\ \hline
    Abnormality & (O) & $\mathbf{41.7 (0.3)}$ & $0.0 (0.0)^{**}$ & $\mathbf{41.7 (0.1)}$ & $0.0 (0.0)^{**}$ \\
    & (C) & $\mathbf{71.5 (0.3)}$ & $43.1 (0.7)^{**}$ & $\mathbf{72.8 (0.6)}$ & $44.7 (0.2)^{**}$  \\ 

    Presence & (C) & $77.7 (0.1)$ & $\mathbf{81.6 (0.3)}^{**}$ & $77.6 (0.4)$ & $\mathbf{81.4 (0.7)}^{**}$  \\ 

    View & (O) & $99.7 (0.0)$ & $99.8 (0.0)$ & $99.6 (0.0)$ & $99.7 (0.1)^{*}$ \\ 
    & (C) & $98.8 (0.1)$ & $\mathbf{99.3 (0.2)}^{*}$  & $99.1 (0.2)$ & $98.9 (0.2)$ \\ 

    Location & (O) & $61.6 (0.4)$ & $\mathbf{64.5 (0.4)^{**}}$ & $61.6 (0.6)$ & $\mathbf{64.5 (0.3)}^{**}$ \\ 

    Level & (O) & $59.5 (0.4)$ & $59.5 (0.7)$ & $60.4 (0.4)$ & $59.9 (1.0)$ \\

    Type & (O) & $60.6 (0.8)$ & $58.8 (0.7)$ & $58.5 (1.0)$ & $59.1 (1.1)$ \\ 
    \hline
\end{tabular}
\end{table}







\section{Expert Model Metrics}\label{app:expert-metrics}
\begin{table}[]
\caption{Quantitative evaluation of the expert model for disease diagnosis (DenseNet121) on $20\%$ of MIMIC-CXR. The AUC performance is reported. 
\url{https://github.com/mlmed/torchxrayvision/blob/master/BENCHMARKS.md}} 
\resizebox{\columnwidth}{!}{%
\begin{tabular}{ccccccccccc}
\hline
Atelectasis &
  Cardiomegaly &
  Consolidation &
  Edema &
  \begin{tabular}[c]{@{}c@{}}Enlarged \\ Cardiomediastinum\end{tabular} &
  Fracture &
  Lung Lesion &
  Lung Opacity &
  Effusion &
  Pneumonia &
  Pneumothorax \\ \hline
0.88 &
  0.88 &
  0.91 &
  0.92 &
  0.84 &
  0.74 &
  0.82 &
  0.86 &
  0.92 &
  0.82 &
  0.81 \\ \hline
\end{tabular}%
}
\end{table}

\end{document}